\documentclass[11pt]{article}
\usepackage[margin=1in]{geometry}
\usepackage{amsmath,amssymb}
\usepackage{graphicx}
\usepackage{booktabs}
\usepackage[unicode]{hyperref}
\usepackage{microtype}
\usepackage[numbers,sort&compress]{natbib}
\usepackage{subcaption}
\usepackage{float}
\title{On the Semantic and Syntactic Information Encoded in Proto-Tokens for One-Step Text Reconstruction}
\author{Ivan Bondarenko\\\texttt{i.bondarenko@g.nsu.ru}\and Egor Palkin\\\texttt{e.palkin1@g.nsu.ru}\and Fedor Tikunov\\\texttt{f.tikunov@g.nsu.ru}}
\date{}

\begin{document}
\maketitle

\begin{abstract}
Autoregressive large language models (LLMs) generate text token-by-token, requiring $n$ forward passes to produce a sequence of length $n$. Recent work, \emph{Exploring the Latent Capacity of LLMs for One-Step Text Reconstruction} (Mezentsev and Oseledets), shows that frozen LLMs can reconstruct hundreds of tokens from only two learned \emph{proto-tokens} in a single forward pass, suggesting a path beyond the autoregressive paradigm.
In this paper, we study what information these proto-tokens encode and how they behave under reconstruction and controlled constraints.
We perform a series of experiments aimed at disentangling semantic and syntactic content in the two proto-tokens, analyzing stability properties of the $e$-token, and visualizing attention patterns to the $e$-token during reconstruction.
Finally, we test two regularization schemes for ``imposing'' semantic structure on the $e$-token using teacher embeddings, including an anchor-based loss and a relational distillation objective.
Our results indicate that the $m$-token tends to capture semantic information more strongly than the $e$-token under standard optimization; anchor-based constraints trade off sharply with reconstruction accuracy; and relational distillation can transfer batch-level semantic relations into the proto-token space without sacrificing reconstruction quality, supporting the feasibility of future non-autoregressive seq2seq systems that predict proto-tokens as an intermediate representation.
\end{abstract}

\section{Introduction}
Autoregressive LLMs generate text in a left-to-right fashion by predicting the next token conditioned on the previously generated context.
This approach has a fundamental computational limitation: generating a sequence of length $n$ typically requires $n$ forward passes of the model.

Non-autoregressive (NAR) text generation aims to reduce this sequential bottleneck by generating multiple tokens (or the whole sequence) in parallel, typically at the cost of additional modeling assumptions, iterative refinement, or distillation from autoregressive teachers.
Recent NAR approaches and analyses provide a broad view of the design space and its trade-offs \citep{nar_methods_survey}.

Recent work, \emph{Exploring the Latent Capacity of LLMs for One-Step Text Reconstruction} (Mezentsev and Oseledets), demonstrates that frozen LLMs can reconstruct hundreds of tokens from only two learned proto-tokens in a single forward pass.
This observation motivates moving beyond the purely autoregressive paradigm.

We view the frozen-LLM plus learned proto-token interface as a complementary perspective on NAR generation: instead of training a fully new NAR model end-to-end, one can (i) keep a strong pretrained decoder frozen, (ii) predict a small set of continuous proto-tokens, and (iii) decode an entire sequence in one pass.
Moreover, our experiments explore explicitly distillation-style constraints (relational distillation) that transfer semantic structure into the proto-token space, connecting this approach to common NAR training recipes \citep{nar_methods_survey}.

Given such a reconstruction mechanism, a natural question arises: can proto-tokens be used not only for reconstruction, but also for \emph{controlled} sequence continuation?
To answer this question, we must understand what information the proto-tokens ($e$ and $m$) contain and how they behave during generation.

In this work, we perform the following experimental study of proto-tokens:
\begin{enumerate}
  \item Semantic content encoded in proto-tokens.
  \item Syntactic content encoded in proto-tokens.
  \item Stability of the $e$-token under optimization and constraints.
  \item Attention visualization to the $e$-token during reconstruction.
  \item Experiments on imposing semantic structure on the $e$-token using a teacher vector.
\end{enumerate}

\section{Related Work}
Our work is most directly influenced by \citet{mezentsev2025onestep} (``Exploring the Latent Capacity of LLMs for One-Step Text Reconstruction'').
That paper shows that, by feeding two trainable proto-tokens into a frozen LLM and optimizing only their input embeddings, one can reconstruct sequences hundreds of tokens in a single forward pass.
Proto-token embeddings are learned by minimizing a cross-entropy reconstruction objective.

For broader context, we refer to recent work on non-autoregressive generation methods and their training strategies, including distillation-based objectives \citep{nar_methods_survey}.

In particular, the input to the frozen LLM can be viewed as a sequence of embeddings where two proto-token embeddings are placed first, followed by a fixed number of repeated $m$ embeddings:
\begin{equation}
  X = \left[ e,\; \underbrace{m,\; m,\; \ldots,\; m}_{T-1} \right].
\end{equation}

Training optimizes only the proto-token embeddings while the model weights remain frozen. Reconstruction is learned with the standard cross-entropy objective:
\begin{equation}
  \mathcal{L}_{\mathrm{CE}}
  = -\frac{1}{T} \sum_{i=1}^{T} \log p_{\theta}(t_i \mid X_{<i})
  = -\frac{1}{T} \sum_{i=1}^{T}
  \log\left( \frac{\exp(z_{i-1,t_i})}{\sum_{c \in V} \exp(z_{i-1,c})} \right).
\end{equation}
Here $z_{i-1,\cdot}$ are the model logits at position $(i-1)$, $V$ is the vocabulary, and $t_i$ is the $i$-th target token.

We extend this direction by analyzing what the two proto-tokens encode.
In particular, we provide evidence that $e$ and $m$ contain both semantic and syntactic information, though to different degrees.
These observations suggest the feasibility of a stable non-autoregressive seq2seq architecture in which a separate model predicts proto-tokens from past context, and a frozen LLM decodes them into continuations (e.g., for question answering).

\section{Method}
\subsection{Proto-token optimization for reconstruction}
We first describe how we obtain proto-token embeddings for one-step reconstruction.
Let the target token sequence be $T = [t_1, \ldots, t_T]$ and let $d$ be the hidden size of the frozen causal LLM.
For each example, we optimize two learnable vectors $e, m \in \mathbb{R}^d$ and build a proto-token-only input embedding sequence
\begin{equation}
  X = [e, m, m, \ldots, m] \in \mathbb{R}^{T \times d},
\end{equation}
where $m$ is repeated $(T-1)$ times.
We feed $X$ to the frozen causal LLM with the standard causal attention mask and obtain logits $Z \in \mathbb{R}^{T \times |V|}$.
Following the next-token prediction convention, the logits at position $(i-1)$ are used to predict token $t_i$.
We optimize only $e$ and $m$ using AdamW \citep{loshchilov2019adamw} until the reconstruction token accuracy reaches a threshold (e.g., $0.9$) or a maximum number of iterations is reached.

\subsection{Semantic augmentation experiment}
To probe semantic content in proto-token representations, for each text in a subset of the dataset we generate $6$ lexical and $9$ semantic augmentations.
For each variant, we optimize proto-token embeddings using the procedure above and then visualize the resulting representations using t-SNE \citep{vandermaaten2008tsne}.
The core hypothesis is: if proto-tokens encode semantic information, then t-SNE projections of proto-tokens for paraphrases of the same example should cluster together, even when surface form differs \citep{vandermaaten2008tsne}.

\paragraph{Lexical augmentations.}
Lexical augmentations are based on introducing typos, including:
(i) random case flips; (ii) frequent orthographic errors; (iii) common keyboard typing errors; and (iv) random deletion/insertion/repetition/swaps of characters.

\paragraph{Semantic augmentations.}
Semantic augmentations are paraphrases generated by an LLM (Qwen3-4B).

\subsection{Imposing semantic structure on the $e$-token}
Because the semantic clustering for the $e$-token was weak in the experiment above, we hypothesize that non-autoregressive optimization admits many good solutions for $(e,m)$, and random initialization can lead to $e$ vectors that do not preserve semantic relations across examples.
This motivates adding additional constraints to encourage semantic structure.

\paragraph{Terminology.}
Throughout the paper we use the terms \emph{proto-tokens}, \emph{$e$-token}, and \emph{$m$-token} consistently.
We refer to the decoding setting studied here as \emph{one-step reconstruction} (not ``one-step generation'').

We evaluate two regularization schemes:
\begin{enumerate}
  \item \textbf{Anchor loss.} We add a term that encourages $e_i$ to stay close to a teacher embedding $t_i$ (obtained from Qwen3-Embedding-8B). We also initialize $e_i$ as $t_i$:
  \begin{equation}
    \mathcal{L}_{\mathrm{anchor}} = 1 - \cos(e_i, t_i).
  \end{equation}
  We combine the reconstruction loss with optional regularizers:
  \begin{equation}
    \mathcal{L}
    =
    \mathcal{L}_{\mathrm{CE}}
    + \lambda_{\mathrm{anchor}}\,\mathcal{L}_{\mathrm{anchor}}
    + \lambda_{\mathrm{rel}}\,\mathcal{L}_{\mathrm{rel}}.
  \end{equation}

  The anchor term is weighted by a separate hyperparameter in the global loss.
  \item \textbf{Relational distillation.} We penalize mismatch between pairwise similarity matrices (using MSE/Huber) computed from teacher embeddings $t_i$ and student embeddings $e_i$ within a batch:
  \begin{equation}
    \mathcal{L}_{\mathrm{rel}}
    =
    \frac{1}{B(B-1)} \sum_{i \ne j} \ell\left(S^{(E)}_{ij},\, S^{(T)}_{ij}\right),
  \end{equation}
  where $B$ is the batch size, $S^{(T)}_{ij} = \cos(t_i, t_j)$ (teacher) and $S^{(E)}_{ij} = \cos(e_i, e_j)$ (student), and $\ell(a,b)=(a-b)^2$ (MSE) or $\ell(a,b)=\mathrm{Huber}(a-b)$.
\end{enumerate}

\subsection{Syntactic experiment}
To probe syntactic content, we generate seven classes of sentences using an LLM and context-free grammars:
(1) simple, (2) complex, (3) simple interrogative, (4) complex interrogative, (5) simple imperative, (6) complex imperative, and (7) one-clause (single-member) sentences.
We then test whether t-SNE projections of proto-token representations cluster by syntactic class.

\subsection{Attention visualization}
We also visualize attention weights to the $e$-token across layers and heads.
Because of causal masking, the $e$-token does not attend to later tokens; thus we focus on how other positions attend \emph{to} $e$ during reconstruction.

\section{Experiments and Results}
\subsection{Attention visualization to the $e$-token}
We visualize attention weights \emph{to} the $e$-token during one-step reconstruction.
Because of causal masking, the $e$-token does not attend to later positions; thus we focus on how other tokens attend to $e$ across layers and heads.
For each example, we first obtain proto-tokens that reconstruct the target sequence well, then run the frozen model forward and extract attention maps.

\begin{figure}[H]
  \centering
  \begin{subfigure}{0.48\linewidth}
    \centering
    \includegraphics[width=\linewidth]{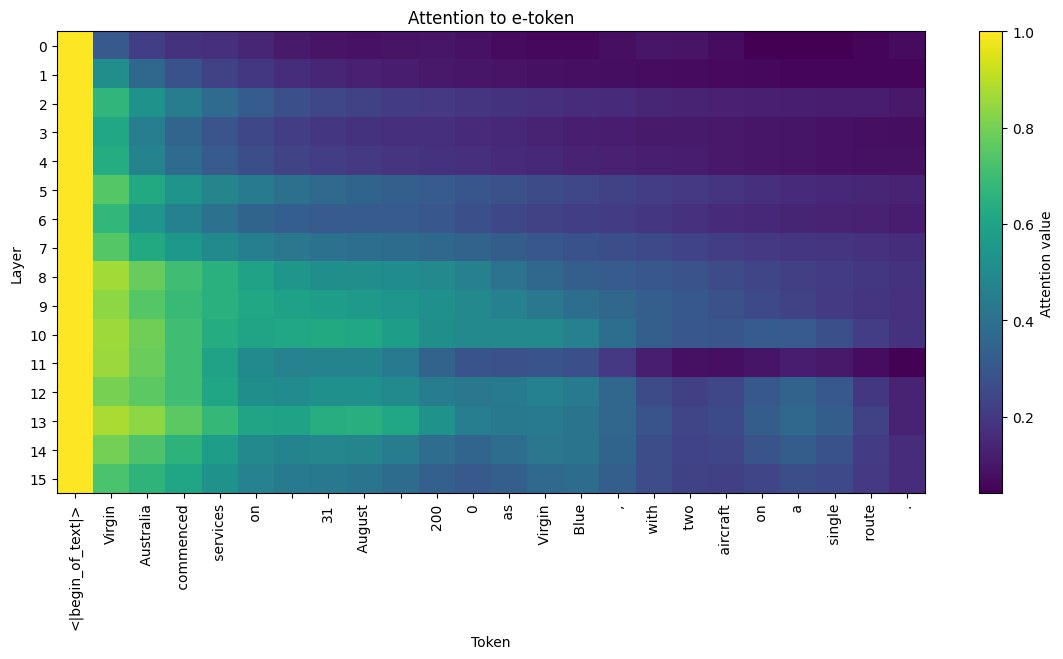}
    \subcaption{Example 1.}
  \end{subfigure}\hfill
  \begin{subfigure}{0.48\linewidth}
    \centering
    \includegraphics[width=\linewidth]{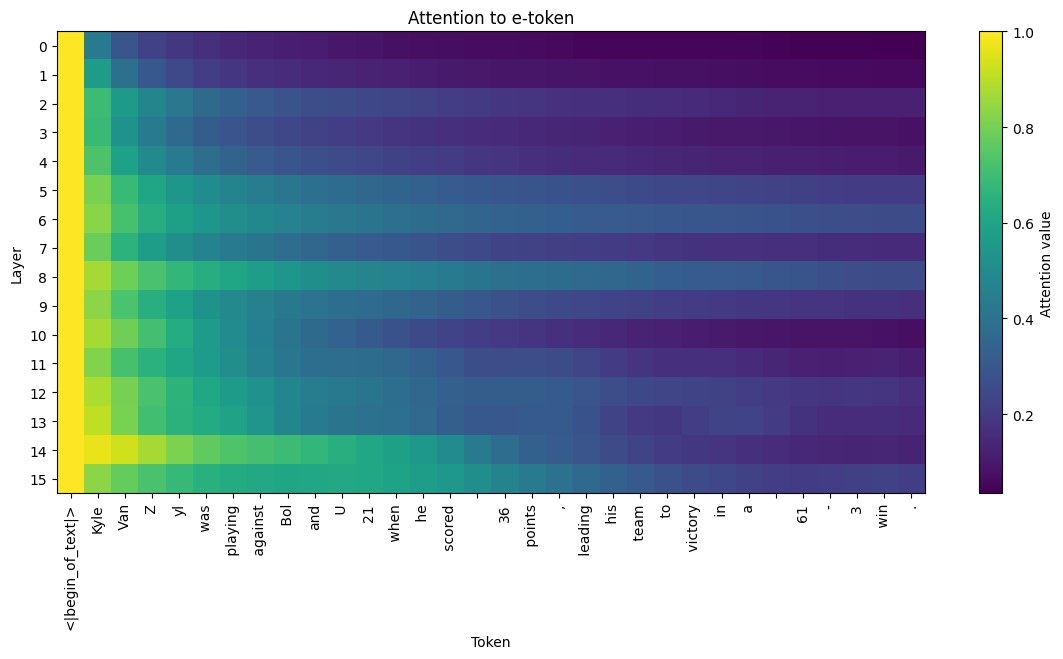}
    \subcaption{Example 2.}
  \end{subfigure}
  \caption{Attention to the $e$-token averaged over heads (examples).}
  \label{fig:attn_mean_over_heads}
\end{figure}

\paragraph{Mean attention over heads.}
Figure~\ref{fig:attn_mean_over_heads} shows examples of attention to the $e$-token averaged over heads for different layers.

\paragraph{Layer-wise mean attention heatmaps.}
We also plot heatmaps of attention to the $e$-token for selected layers, averaged over heads.

\begin{figure}[H]
  \centering
  \begin{subfigure}{0.48\linewidth}
    \centering
    \includegraphics[width=\linewidth]{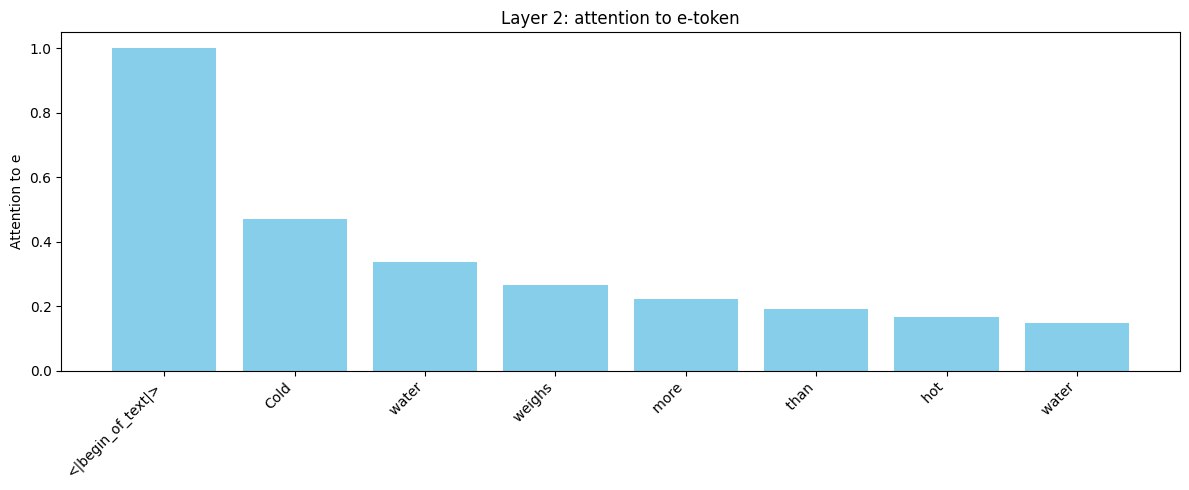}
    \subcaption{Layer 0.}
  \end{subfigure}\hfill
  \begin{subfigure}{0.48\linewidth}
    \centering
    \includegraphics[width=\linewidth]{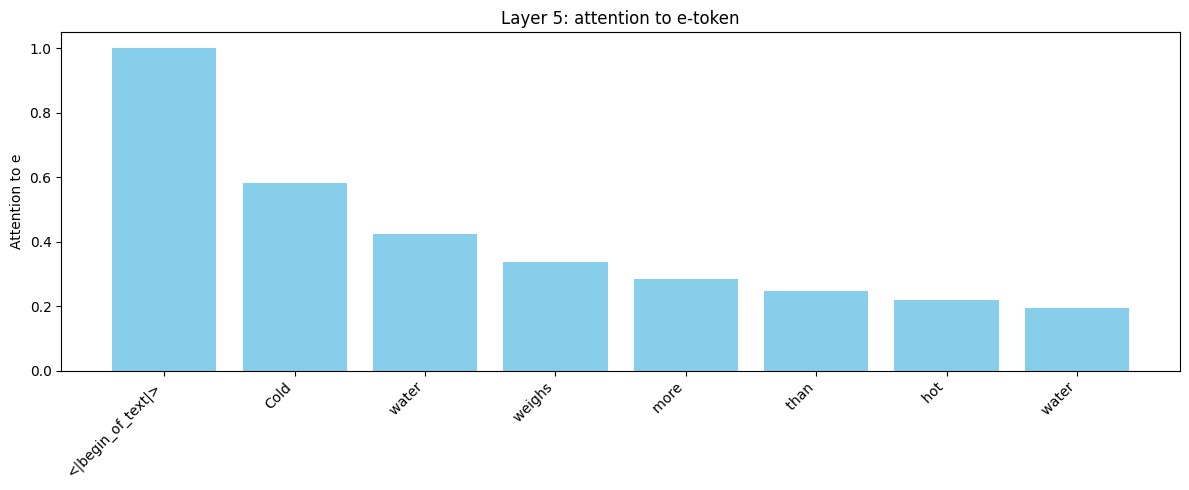}
    \subcaption{Layer 4.}
  \end{subfigure}
  \vspace{0.6em}

  \begin{subfigure}{0.48\linewidth}
    \centering
    \includegraphics[width=\linewidth]{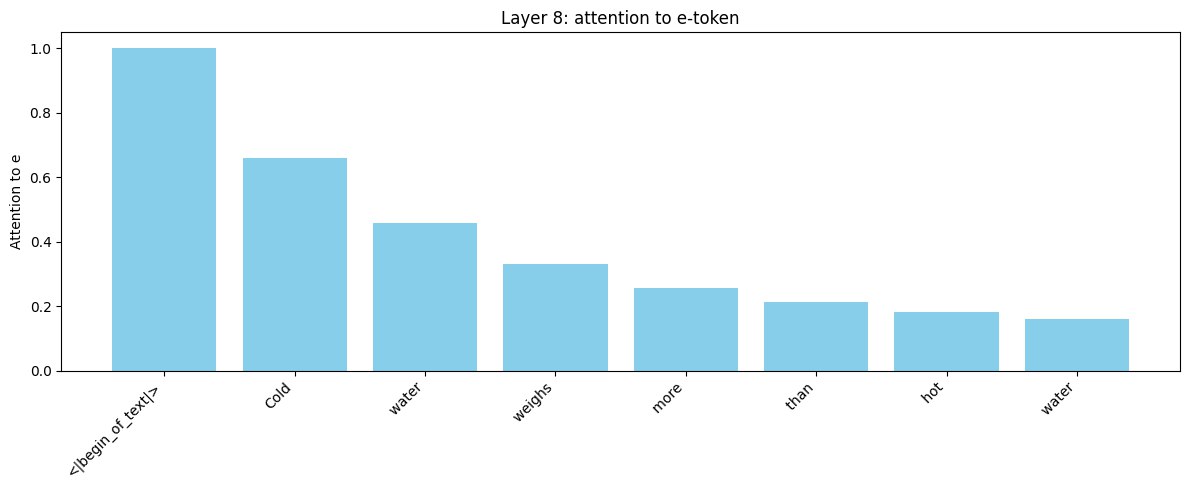}
    \subcaption{Layer 8.}
  \end{subfigure}\hfill
  \begin{subfigure}{0.48\linewidth}
    \centering
    \includegraphics[width=\linewidth]{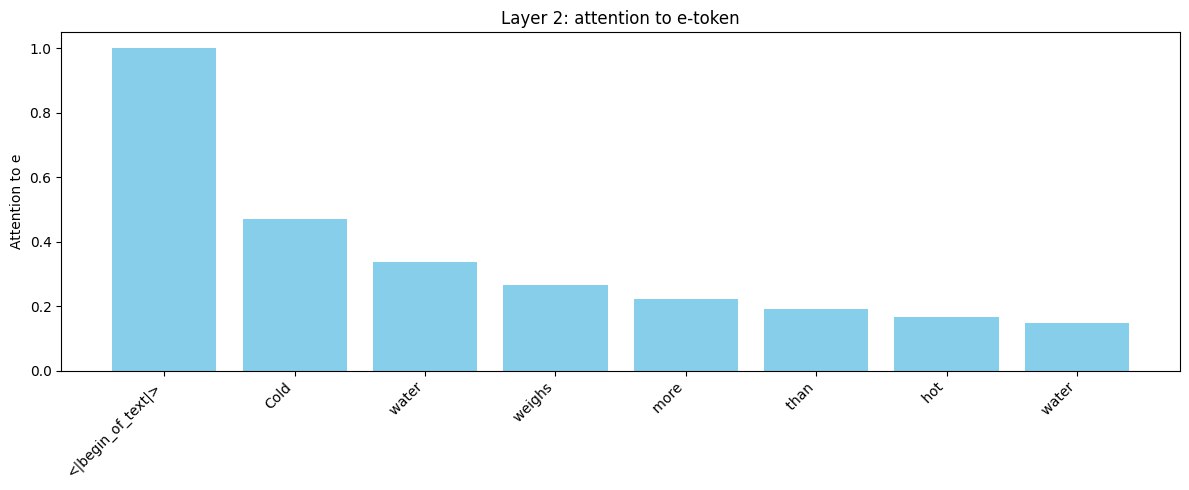}
    \subcaption{Layer 12.}
  \end{subfigure}
  \caption{Layer-wise heatmaps of attention to the $e$-token (averaged over heads).}
  \label{fig:attn_layer_heatmaps}
\end{figure}

\begin{figure}[H]
  \centering
  \begin{subfigure}{0.48\linewidth}
    \centering
    \includegraphics[width=\linewidth]{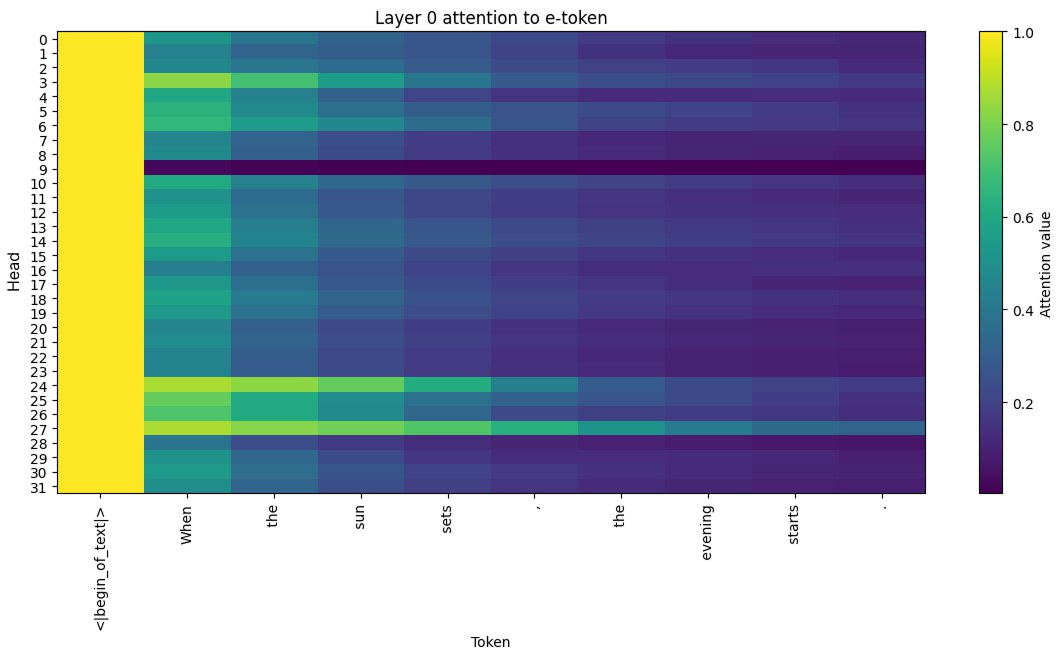}
    \subcaption{Head example 1.}
  \end{subfigure}\hfill
  \begin{subfigure}{0.48\linewidth}
    \centering
    \includegraphics[width=\linewidth]{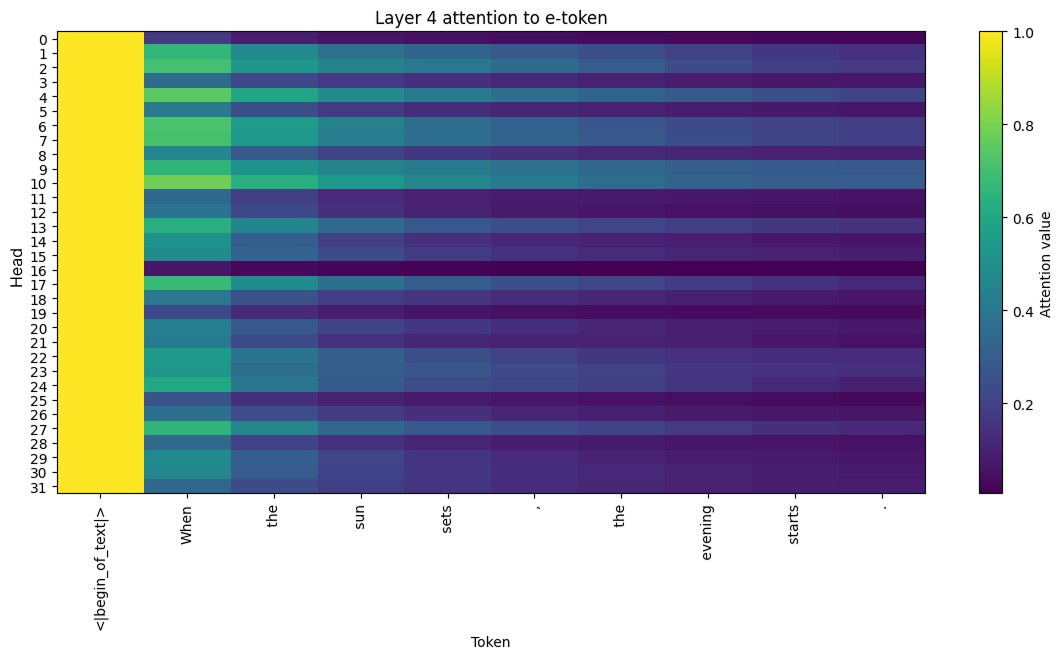}
    \subcaption{Head example 2.}
  \end{subfigure}
  \vspace{0.6em}

  \begin{subfigure}{0.48\linewidth}
    \centering
    \includegraphics[width=\linewidth]{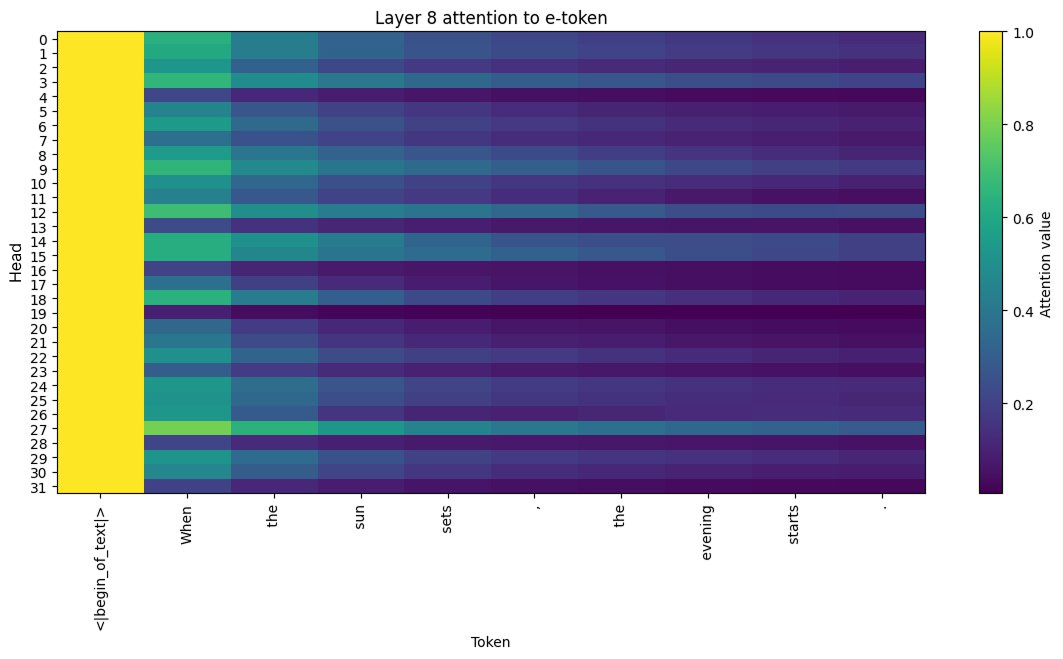}
    \subcaption{Head example 3.}
  \end{subfigure}\hfill
  \begin{subfigure}{0.48\linewidth}
    \centering
    \includegraphics[width=\linewidth]{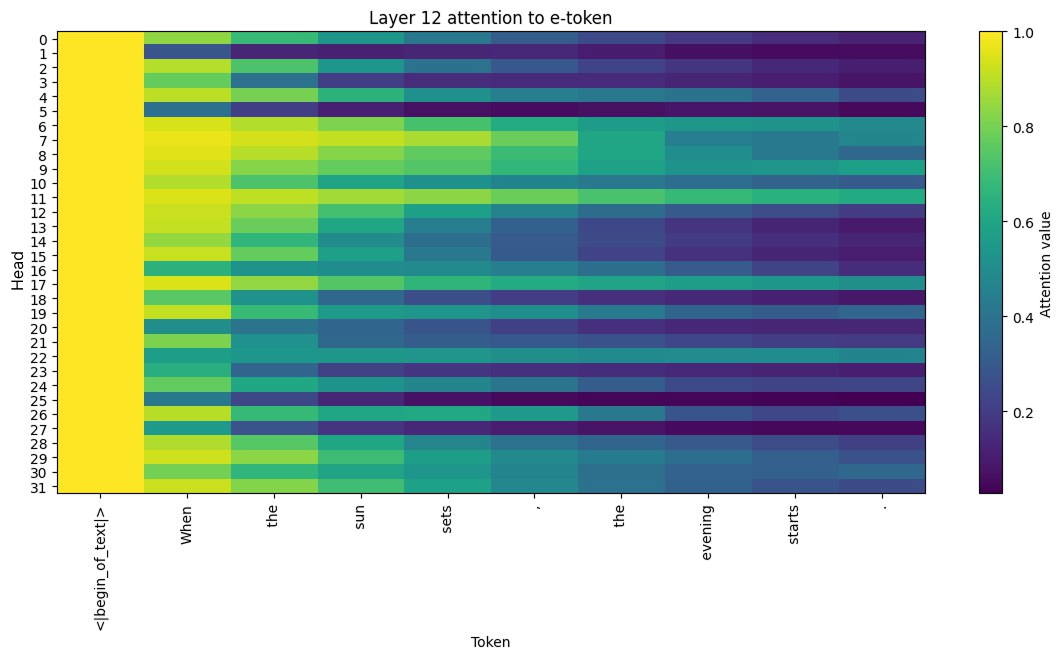}
    \subcaption{Head example 4.}
  \end{subfigure}
  \caption{Head-level attention to the $e$-token for selected layers/heads (examples).}
  \label{fig:attn_headlevel}
\end{figure}

\paragraph{Head-level attention distributions.}
Finally, we visualize attention to the $e$-token for individual heads in selected layers (Figure~\ref{fig:attn_headlevel}).

Overall, we observe that attention to the $e$-token is typically strongest for early positions and tends to decay for later tokens; the exact pattern varies by layer and head.

\subsection{Robustness to additive noise in proto-tokens}
We study how additive noise affects reconstruction quality when perturbing the optimized proto-token embeddings, focusing on the stability of the reconstruction process under embedding-level disturbances.
Given optimized proto-tokens $(e,m)$ for a text example, we add noise to the $e$-token and measure reconstruction token accuracy as a function of noise intensity and distribution.

\paragraph{Noise distributions.}
We consider four types of noise:
\begin{enumerate}
  \item Gaussian: $\varepsilon \sim \mathcal{N}(0,\sigma^2)$.
  \item Uniform: $\varepsilon \sim \mathcal{U}(a,b)$.
  \item Exponential: $\varepsilon \sim \mathrm{Exp}(\lambda)$.
  \item Sinusoidal: $\varepsilon = \sin(\omega i + \phi)$ (structured perturbation).
\end{enumerate}

\paragraph{Normalization.}
We scale the noise to control its magnitude relative to the original embedding norm:
\begin{equation}
  \tilde{\varepsilon}
  =
  \frac{\varepsilon}{\lVert \varepsilon \rVert_2}\cdot \alpha \cdot \lVert e \rVert_2,
\end{equation}
where $\alpha \in \{0.0, 0.05, 0.1, 0.2, 0.5, 1.0\}$ controls intensity.

\begin{figure}[H]
  \centering
  \includegraphics[width=0.7\linewidth]{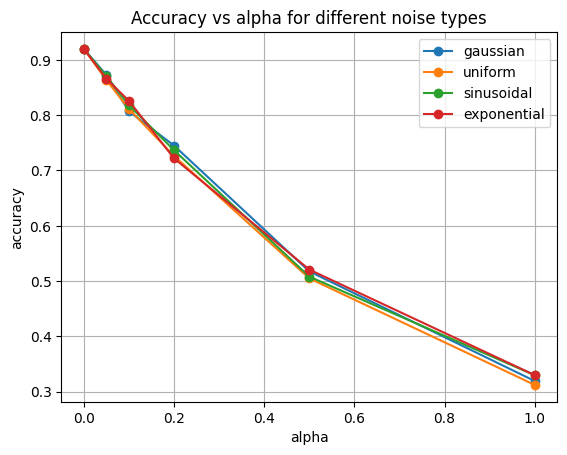}
  \caption{Reconstruction accuracy as a function of noise intensity $\alpha$ for different noise types.}
  \label{fig:noise_effect}
\end{figure}

As expected, reconstruction quality degrades as the noise level increases; different noise distributions exhibit similar trends in our setup.

\subsection{Models and data}
\paragraph{Models.}
We use Llama-3.2-1B to obtain optimized proto-token embeddings, Qwen3-4B to generate paraphrases, and Qwen3-Embedding-8B to produce teacher vectors.

\paragraph{Data.}
For the semantic experiment, we use the databricks-dolly-15k dataset of instruction-following pairs \citep{dolly15k_hf}.
We generate paraphrases via Qwen3-4B and typo-based variants via the \texttt{augmentex} library \citep{augmentex}.
For semantic-imposition and attention-visualization experiments, we also use Dolly.
For the syntactic experiment, we use fully synthetic data generated by Qwen3-4B and context-free grammars; the grammar vocabulary is built from the 500 most frequent words in the Brown corpus categories ``news'', ``editorial'', ``learned'', and ``reviews'' \citep{brown_corpus}, with POS tags assigned via WordNet matches \citep{miller1995wordnet}.

\paragraph{Optimization details.}
Proto-token embeddings are optimized with AdamW \citep{loshchilov2019adamw} using learning rate $0.01$, $\beta_1=\beta_2=0.9$, and weight decay $0.01$.
We use $\beta_1=\beta_2=0.9$ following preliminary tuning.
We optimize each example until token accuracy reaches $0.9$ or for at most 2000 iterations.
Embeddings are initialized from a standard normal distribution unless otherwise stated.

\subsection{Semantic content: $e$ vs.\ $m$}
From t-SNE visualizations, we observe that the $m$-token embeddings exhibit stronger semantic clustering than $e$ under standard optimization.
This suggests that $m$ carries more semantic information about the sentence content, while $e$ may encode other factors (or be less identifiable across solutions).

\begin{figure}[H]
  \centering
  \includegraphics[width=0.65\linewidth]{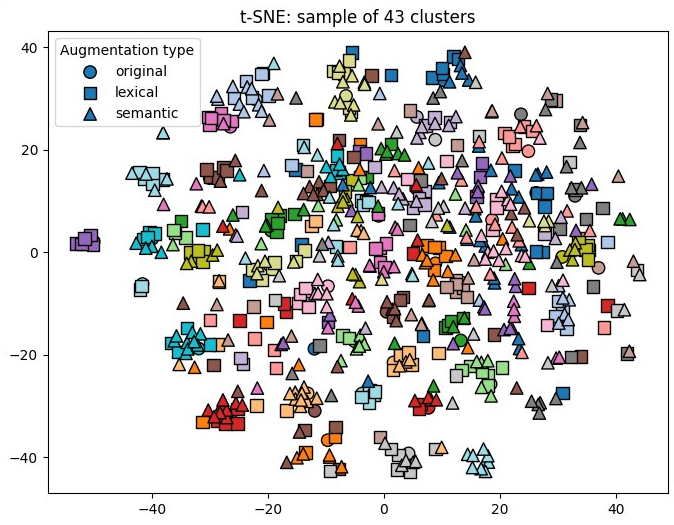}
  \caption{t-SNE visualization of optimized $e$-token embeddings for original, lexical, and semantic augmentations.}
  \label{fig:tsne_e}
\end{figure}

\begin{figure}[H]
  \centering
  \includegraphics[width=0.65\linewidth]{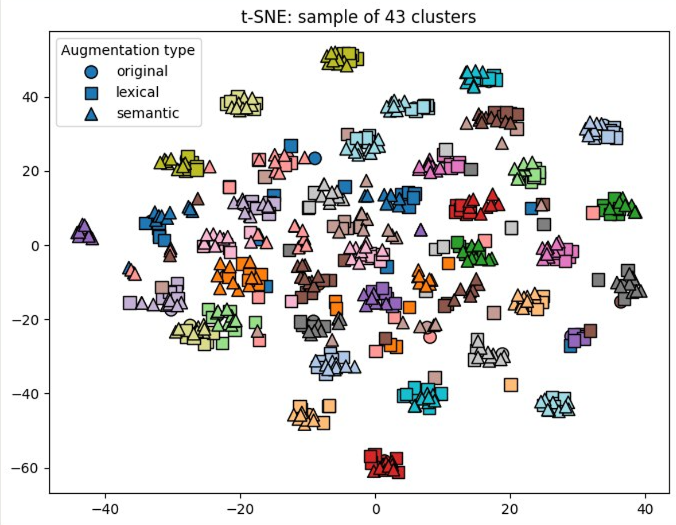}
  \caption{t-SNE visualization of optimized $m$-token embeddings for original, lexical, and semantic augmentations.}
  \label{fig:tsne_m}
\end{figure}

\subsection{Imposing semantics on $e$}
\paragraph{Anchor loss.}
With a small anchor weight (e.g., $0.02$ multiplying the anchor term in the global loss), reconstruction can remain mostly successful.
However, the final cosine similarity between $e$ and the teacher vector remains close to zero in our runs.
Increasing the anchor weight (e.g., to $0.5$) substantially degrades reconstruction accuracy while increasing the final similarity.
Overall, these results do not support the hypothesis that $e$ can be forced to match an off-the-shelf sentence embedding while preserving reconstruction quality.

\begin{figure}[H]
  \centering
  \includegraphics[width=0.65\linewidth]{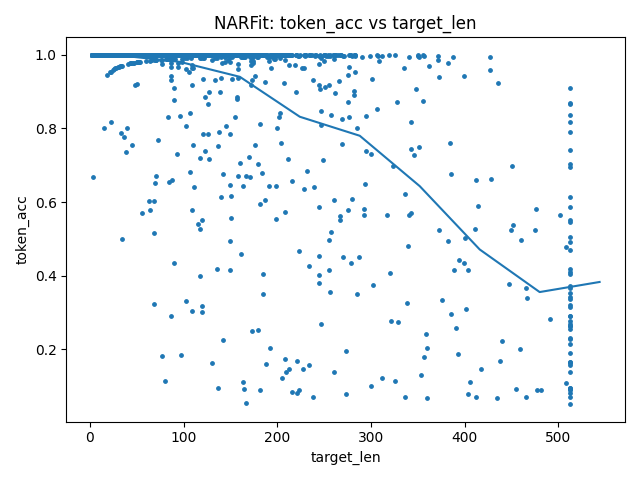}
  \caption{Reconstruction token accuracy vs.\ target length under anchor regularization with a small weight.}
  \label{fig:anchor_acc_small}
\end{figure}

\begin{figure}[H]
  \centering
  \includegraphics[width=0.65\linewidth]{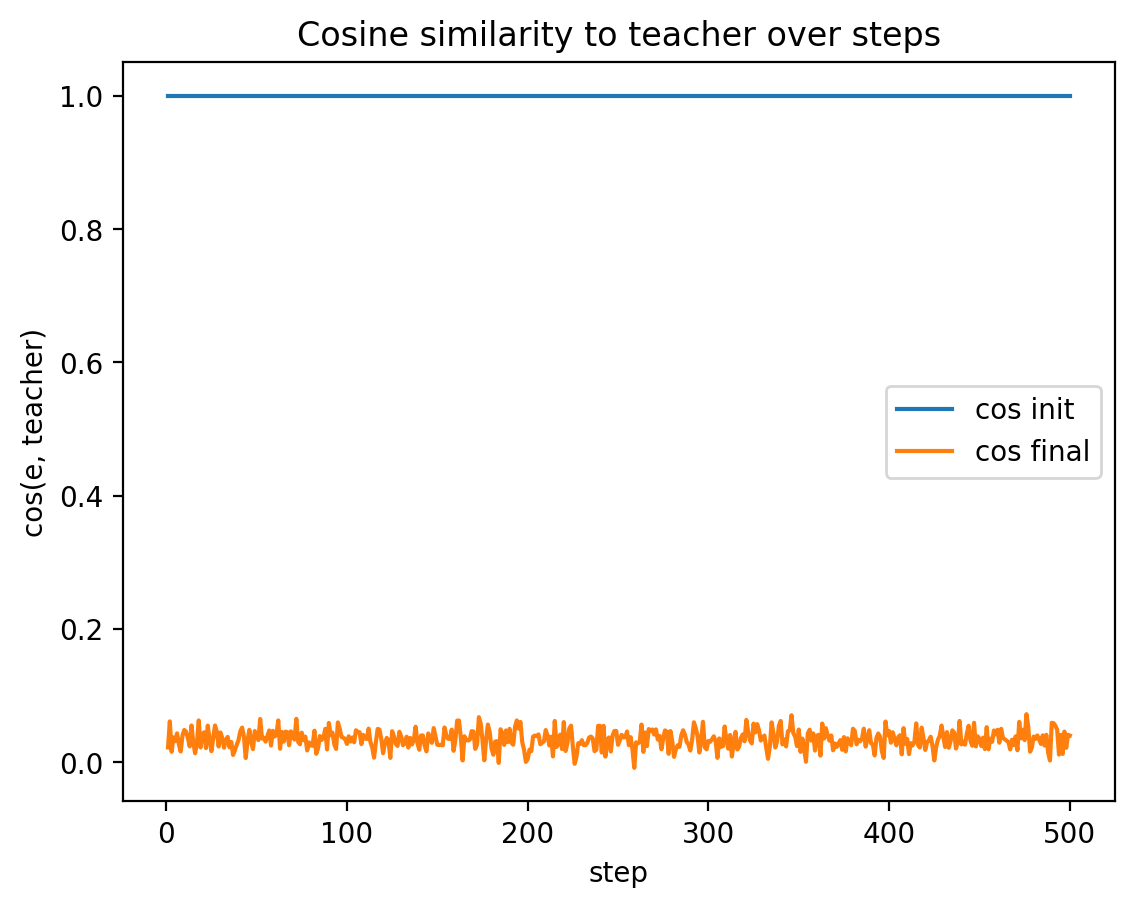}
  \caption{Cosine similarity between $e$ and the teacher embedding over optimization steps (anchor).}
  \label{fig:anchor_cos}
\end{figure}

\begin{figure}[H]
  \centering
  \includegraphics[width=0.65\linewidth]{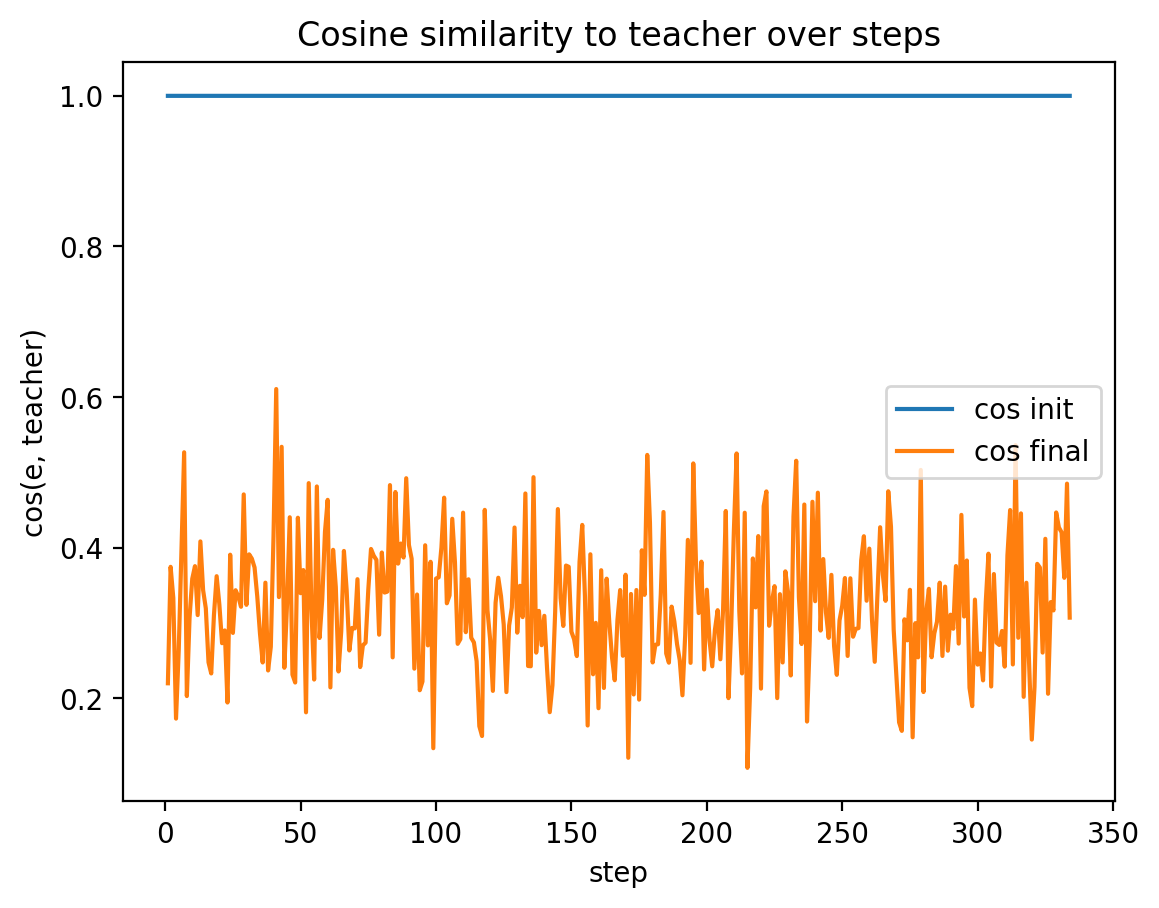}
  \caption{Cosine similarity between $e$ and the teacher embedding over optimization steps for a larger anchor weight.}
  \label{fig:anchor_cos_large}
\end{figure}

\begin{figure}[H]
  \centering
  \includegraphics[width=0.65\linewidth]{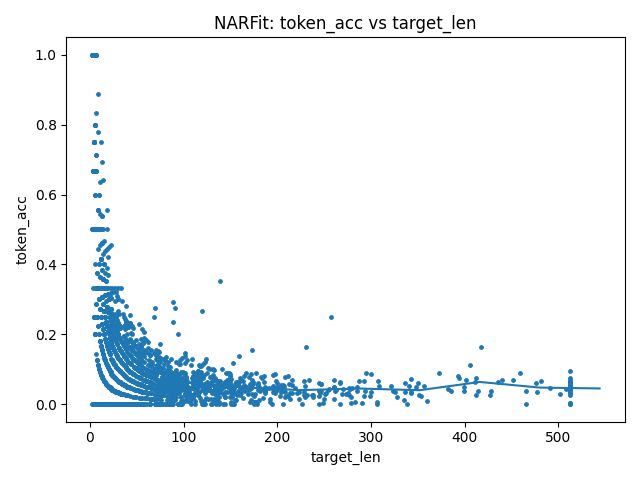}
  \caption{Token accuracy vs.\ target length for a larger anchor weight, showing degraded reconstruction.}
  \label{fig:anchor_acc_large}
\end{figure}

\paragraph{Relational distillation.}
Using relational distillation, we find that even without constraints there is weak semantic structure in $e$-tokens, but it becomes more apparent when using a shared $m$-token.
Importantly, adding the relational constraint preserves the ability to reconstruct text accurately while significantly improving correlation between teacher and student similarity matrices across batches (batch size 6).
This indicates that relational distillation can transfer semantic relations into the proto-token space without sacrificing reconstruction.

\begin{figure}[H]
  \centering
  \begin{subfigure}{0.48\linewidth}
    \centering
    \includegraphics[width=\linewidth]{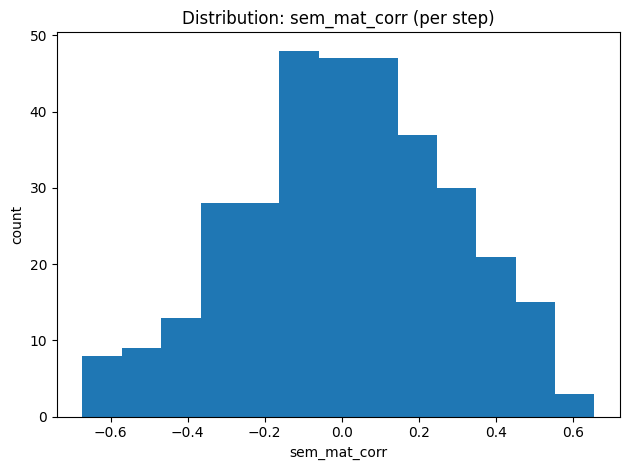}
    \subcaption{No shared $m$.}
  \end{subfigure}\hfill
  \begin{subfigure}{0.48\linewidth}
    \centering
    \includegraphics[width=\linewidth]{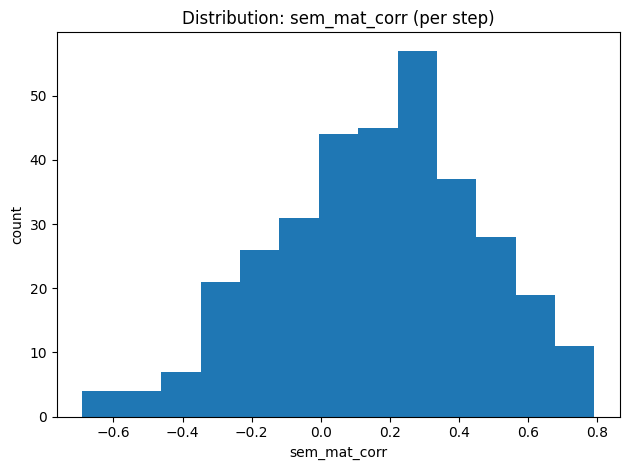}
    \subcaption{Shared $m$.}
  \end{subfigure}
  \caption{Distributions of semantic-matrix correlation per step for relational distillation, with and without a shared $m$-token.}
  \label{fig:rel_semcorr}
\end{figure}

\begin{figure}[H]
  \centering
  \begin{subfigure}{0.48\linewidth}
    \centering
    \includegraphics[width=\linewidth]{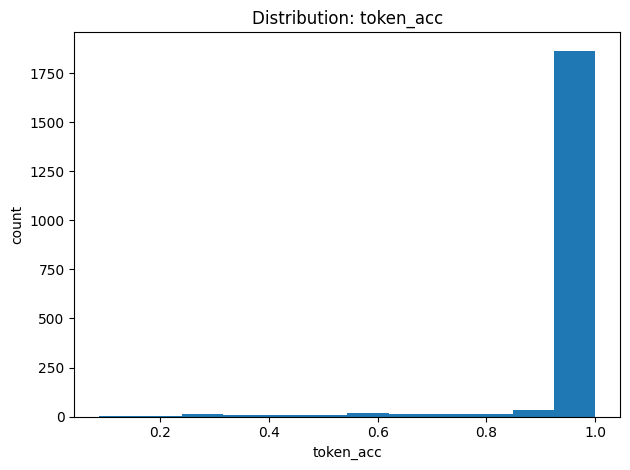}
    \subcaption{Without constraint.}
  \end{subfigure}\hfill
  \begin{subfigure}{0.48\linewidth}
    \centering
    \includegraphics[width=\linewidth]{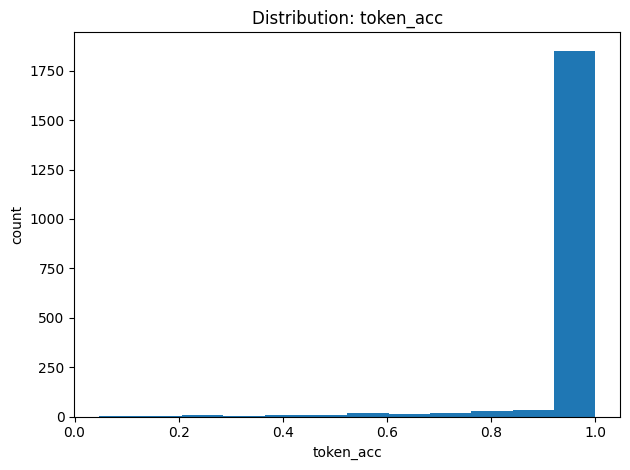}
    \subcaption{With constraint.}
  \end{subfigure}
  \caption{Token-accuracy distributions with and without the relational distillation constraint.}
  \label{fig:rel_acc}
\end{figure}

\begin{figure}[H]
  \centering
  \begin{subfigure}{0.48\linewidth}
    \centering
    \includegraphics[width=\linewidth]{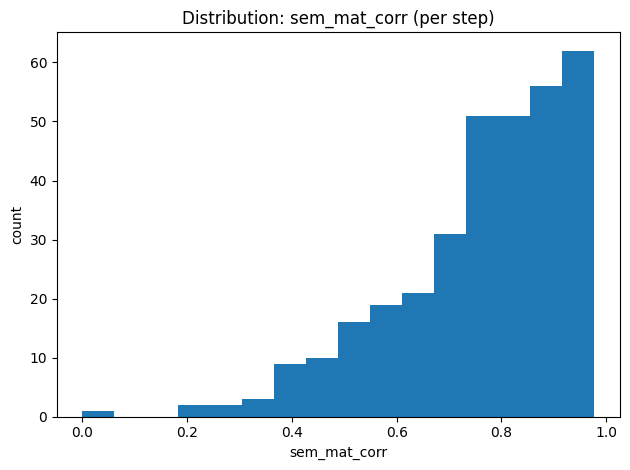}
    \subcaption{With shared $m$.}
  \end{subfigure}\hfill
  \begin{subfigure}{0.48\linewidth}
    \centering
    \includegraphics[width=\linewidth]{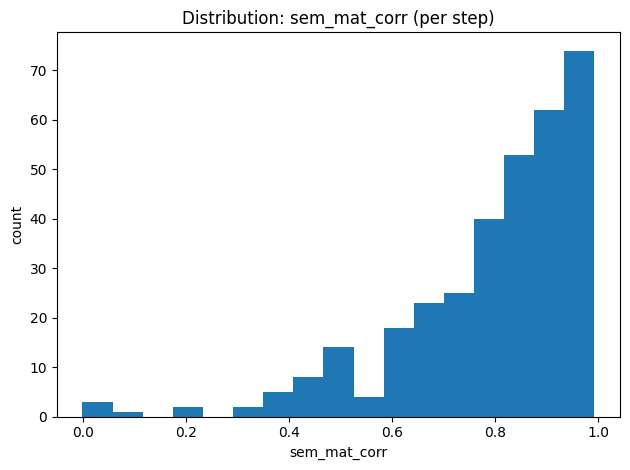}
    \subcaption{Without shared $m$.}
  \end{subfigure}
  \caption{Improved semantic-matrix correlation distributions after applying the relational distillation objective.}
  \label{fig:rel_semcorr_improved}
\end{figure}

\subsection{Syntactic content and attention patterns}
Our syntactic clustering and attention analyses suggest that proto-tokens also encode information correlated with syntactic structure.
These observations support the working hypothesis that, if $m$ is more semantically aligned, then $e$ may carry more syntactic information; we leave a detailed quantitative study as future work.

\section{Discussion and Conclusion}
We studied proto-tokens used for one-step reconstruction with frozen LLMs and investigated what information they encode.
Our experiments indicate that:
\begin{itemize}
  \item Under unconstrained optimization, the $m$-token tends to capture semantic information more strongly than the $e$-token.
  \item Anchor-based regularization that forces $e$ toward a teacher sentence embedding introduces a sharp trade-off: stronger anchoring improves similarity but quickly harms reconstruction accuracy, and weak anchoring does not reliably align $e$ with the teacher space.
  \item Relational distillation is a more promising alternative: it transfers batch-level semantic relations into the proto-token space while preserving high reconstruction performance.
  \item Proto-tokens appear to encode syntactic information as well, and attention patterns suggest that the model actively uses the $e$-token during decoding.
\end{itemize}

These findings support the feasibility of a non-autoregressive seq2seq pipeline in which a lightweight predictor maps context to proto-tokens, and a frozen LLM decodes them into text.
Future work includes (i) building an explicit proto-token predictor for controlled generation, (ii) quantifying syntactic vs.\ semantic disentanglement, and (iii) improving identifiability and stability of proto-token representations across optimization runs.

\bibliographystyle{plainnat}

\clearpage

\appendix

\section{Training and Optimization Details}

\subsection{Prototype Token Initialization}

Prototype token embeddings were initialized from a standard normal distribution.

\subsection{Optimization}

Embeddings were optimized using the AdamW optimizer with the following hyperparameters:
\begin{itemize}
    \item Learning rate: 0.01
    \item $\beta_1 = 0.9$
    \item $\beta_2 = 0.9$
    \item Weight decay: 0.01
\end{itemize}

All other parameters were kept at default values.

\section{Semantic Augmentation Experiment}  

\subsection{Lexical augmentations}

Character-level lexical augmentation was performed using the \texttt{augmentex} library with the following configuration:
\begin{itemize}
    \item Character modification probability: 0.3
    \item Minimum number of augmentations per sentence: 1
    \item Maximum number of augmentations per sentence: 5
    \item Maximum character repetition factor: 3
    \item Keyboard layout for typo simulation: PC
\end{itemize}

\subsection{Semantic augmentations}

Semantic paraphrases were generated using the Qwen3-4B language model. The decoding parameters were:
\begin{itemize}
    \item Temperature: 0.7
    \item Top-p: 0.9
\end{itemize}

\section{Syntactic Augmentation Experiment}

To probe syntactic content, we generated seven sentence classes using an LLM and context-free grammars.

\subsection{Lexical Resources for CFG Generation}
Lexical resources for CFG generation were derived as follows:
\begin{enumerate}
    \item The 500 most frequent words were extracted from the Brown corpus, restricted to the categories \texttt{news}, \texttt{editorial}, \texttt{learned}, and \texttt{reviews}.
    \item Each word was assigned a part of speech using WordNet synsets and POS tagging to ensure grammatical correctness.
\end{enumerate}

\subsection{LLM Generation}
For syntactic sentence generation via the LLM, we used Qwen3-4B. The decoding parameters were:
\begin{itemize}
    \item Temperature: 0.7
    \item Top-p: 0.9
\end{itemize}

\end{document}